\documentclass[11pt]{article}
\usepackage{emnlp2016}
\usepackage[table]{xcolor}
\usepackage{times}
\usepackage{url}
\usepackage{latexsym}
\usepackage{graphicx}
\usepackage{subcaption}
\usepackage{subfig}
\usepackage{pgfplotstable}
\usepackage{adjustbox}
\usepackage{booktabs}
\usepackage{amssymb}
\usepackage{amsmath}
\usepackage{multirow,bigdelim}
\usepackage{latexsym}
\usepackage{paralist}
\usepackage{xspace}
\usepackage{microtype}
\usepackage{relsize}
\usepackage{adjustbox}

\emnlpfinalcopy

\newcommand{\ex}[1]{\textit{#1}\xspace}
\newcommand{\class}[1]{\textsf{#1}\xspace}

\usepackage{titlesec}
\usepackage{lipsum}
\usepackage{booktabs}

\titlespacing{\paragraph}{%
  0pt}{
  0.2\baselineskip}{
  0.5em}

\newcommand{\typeAlign}{\method{LabelEmbed}}

\title{Named Entity Recognition for Novel Types by Transfer Learning}

\author{Lizhen Qu$^{1,2}$, Gabriela Ferraro$^{1,2}$, Liyuan Zhou$^1$,\\
\textbf{Weiwei Hou}$^1$, \textbf{Timothy Baldwin}$^{1,3}$ \\[1ex]
$^1$ DATA61, Australia \\
$^2$ The Australian National University\\
$^3$ The University of Melbourne \\
{\small\{lizhen.qu,gabriela.ferraro,joe.zhou\}@data61.csiro.au}\\
{\small{houvivid2013@gmail.com}},
{\small{tb@ldwin.net}}\\
}
\date{}

\newcommand{\dataset}[1]{\texttt{#1}\xspace}
\newcommand{\Conll}{\dataset{CoNLL}}
\newcommand{\Ib}{\dataset{I2B2}}	
\newcommand{\BBN}{\dataset{BBN}}

\newcommand{\CADEC}{\dataset{CADEC}}

\newcommand{\method}[1]{\textbf{#1}\xspace}
\newcommand{\transDeep}{\method{TransDeepCRF}}
\newcommand{\transInit}{\method{TransInit}}
\newcommand{\embed}{\method{Embed}}
\newcommand{\cca}{\method{CCA}}

\newcommand{\eqnref}[1]{Equation~\eqref{#1}\xspace}
\newcommand{\figref}[1]{Figure~\ref{#1}\xspace}

\newcommand{\secref}[1]{Section~\ref{#1}\xspace}

\begin{document}

\maketitle

\begin{abstract}
  In named entity recognition, we often don't have a large in-domain training
  corpus or a knowledge base with adequate coverage to train a model
  directly. In this paper, we propose a method where, given training
  data in a related domain with similar (but not identical) named entity
  (NE) types and a small amount of in-domain training data, we use transfer learning to learn a domain-specific NE
  model. That is, the novelty in the task setup is that we assume not
  just domain mismatch, but also label mismatch.
\end{abstract}

\section{Introduction}
\label{sec:intro}

There are two main approaches to named entity recognition (NER): (i) build sequence labelling models such as conditional random fields (CRFs)~\cite{lafferty2001conditional} on a large manually-labelled training corpus~\cite{finkel2005incorporating}; and (ii) exploit knowledge bases to recognise mentions of entities in text~\cite{rizzo2012nerd,mendes2011dbpedia}. For many social media-based or security-related applications, however, we cannot assume that we will have access to either of these. An alternative is to have a small amount of in-domain training data and access to large-scale annotated data in a second domain, and perform transfer learning over both the features and label set. This is the problem setting in this paper.

NER of novel named entity (NE) types poses two key challenges. First is the issue of sourcing labelled training data. Handcrafted features play a key role in supervised NER models~\cite{turian2010word}, but if we have only limited training amounts of training data, we will be hampered in our ability to reliably learn feature weights. Second, the absence of target NE types in the source domain makes transfer difficult, as we cannot directly apply a model trained over the source domain to the target domain. \newcite{finCorpus:2015} show that even if the NE label set is identical across domains, large discrepancies in the label distribution can lead to poor performance.

Despite these difficulties, it is possible to transfer knowledge between domains, as related NE types often share lexical and context features. For example, the expressions \ex{give lectures} and \ex{attend tutorials} often occur near mentions of NE types \class{PROFESSOR} and \class{STUDENT}. If only \class{PROFESSOR} is observed in the source domain but we can infer that the two classes are similar, we can leverage the training data to learn an NER model for \class{STUDENT}. In practice, differences between NE classes are often more subtle than this, but if we can infer, for example, that the novel NE type \class{STUDENT} aligns with NE types \class{PERSON} and \class{UNIVERSITY}, we can compose the context features of \class{PERSON} and \class{UNIVERSITY} to induce a model for \class{STUDENT}.

In this paper, we propose a transfer learning-based approach to NER in novel domains with label mismatch over a source domain. We first train an NER model on a large source domain training corpus, and then learn the correlation between the source and target NE types. In the last step, we reuse the model parameters of the second step to initialise a linear-chain CRF and fine tune it to learn domain-specific patterns. We show that our methods achieve up to 160\% improvement in F-score over a strong baseline, based on only 125 target-domain training sentences. 




\section{Related work}
\label{sec:relwork}



The main scenario where transfer learning has been applied 
to NER is domain adaptation \cite{Arnold:08,Maynard:01,Chiticariu:2010}, where it is assumed that the label set $Y$ is the same for both the source and target corpora, and only the domain varies. 
In our case, however, both the domain and the label set differ across datasets.


Similar to our work, \newcite{Kim:2015} use transfer learning to deal with NER data sets with different label distributions. 
They use canonical correlation analysis (CCA) to induce label representations, and reduce the problem to one of domain adaptation.
This supports two different label mappings: (i) to a coarse label set by clustering vector representations of the NE types, which are combined with mention-level predictions over the target domain to train a target domain model; and (ii) between labels based on the $k$ nearest neighbours of each label type, and from this transferring a pre-trained model from the source to the target domain. 
They showed their automatic label mapping strategies attain better results than a manual mapping, with the pre-training approach achieving the best results.
Similar conclusions were reached by \newcite{Yosinski:2014}, who investigated the transferability of features from a deep neural network trained over the ImageNet data set. 
\newcite{Sutton:05} investigated how the target task affects the source task, 
and demonstrated that decoding for transfer is better than no transfer, and joint decoding is better than cascading decoding.

Another way of dealing with a lack of annotated NER data is to use distant supervision by exploiting knowledge bases to recognise mentions of entities \cite{Figer,dong:2015:IJCAI,yosef:2013:hyena,Althobaiti:TACL2015,Yaghoobzadeh:15:emnlp}.
Having a fine-grained entity typology has been shown to improve other tasks such as relation extraction \cite{Figer} and question answering \cite{lee2007fine}. 
Nevertheless, for many social media-based or security-related applications, we don't have access to a high-coverage knowledge base, meaning distant supervision is not appropriate.






\section{Transfer Learning for NER}
\label{sec:transfer_ner}

Our proposed approach \transInit consists of three steps: (1) we train a linear-chain CRF on a large source-domain corpus; (2) we learn the correlation between source NE types and target NE types using a two-layer neural network; and (3) we leverage the neural network to train a CRF for target NE types.

Given a word sequence $\mathbf{x}$ of length $L$, an NER system assigns each word $x_i$ a label $y_i\in \mathcal{Y}$, where the label space $\mathcal{Y}$ includes all observed NE types and a special category \class{O} for words without any NE type.  Let $(\mathbf{x}, \mathbf{y})$ be a sequence of words and their labels. A linear-chain CRF takes the form:
\begin{equation}
\label{eq:crf}
\frac{1}{Z} \prod_{l = 1}^{L}\exp \bigg(\mathbf{W}^f f(y_l, \mathbf{x})
+ \mathbf{W}^g g(y_{l-1}, y_l)  \bigg),
\end{equation} 
where $f(y_l, \mathbf{x})$ is a feature function depending only on $\mathbf{x}$, and the feature function $g(y_{l-1}, y_l)$ captures co-occurrence between adjunct labels. The feature functions are weighted by model parameters $\mathbf{W}$, and $Z$ serves as the partition function for normalisation. 

The source domain model is a linear-chain CRF trained on a labelled source corpus. The co-occurrence of target domain labels is easy to learn due to the small number of parameters ($|Y|^2$). Mostly such information is domain specific so that it is unlikely that the co-occurrence of two source types can be matched to the co-occurrence of the two target types. However the feature functions $f(y_l, \mathbf{x})$ capture valuable information about the textual patterns associated with each source NE type. Without  $g(y_{l-1}, y_l)$, the linear-chain CRF is reduced to a logistic regression (LR) model: 
\begin{equation}
 \sigma(y^*, \mathbf{x}_i; \mathbf{W}^f) = \frac{\exp ( \mathbf{W}^f_{.y^*} f(y^*_i, \mathbf{x_i}) )}{\sum_{y \in \mathcal{Y}} \exp(\mathbf{W}^f_{.y} f(y, \mathbf{x}_i))}.
\end{equation} 

In order to learn the correlation between source and target types, we formulate it as a predictive task by using the unnormalised probability of source types to predict the target types. Due to the simplification discussed above, we are able to extract a linear layer from the source domain, which takes the form
$\mathbf{a}_i = \mathbf{W^s}\mathbf{x}_i$, 
where $\mathbf{W^s}$ denotes the parameters of $f(y_l, \mathbf{x})$ in the source domain model, and each $\mathbf{a}_i$ is the unnormalised probability for each source NE type. Taking $\mathbf{a}_i$ as input, we employ a multi-class LR classifier to predict target types, which is essentially 
 $p(y' | \mathbf{a}) = \sigma(y',\mathbf{a}_i ;  \mathbf{W^t})$, 
where $y'$ is the observed type. From another point of view, the whole architecture is a neural network with two linear layers. 

We do not add any non-linear layers between these two linear layers because we otherwise end up with saturated activation functions. An activation function is saturated if its input values are its max/min values~\cite{DBLP:journals/jmlr/GlorotB10}. Taking $\tanh(x)$ as an example, $\frac{\partial tanh(z)}{\partial z} = 1 - \tanh^2(z)$.  If $z$ is, for example, larger than 2, the corresponding derivative is smaller than 0.08. Assume that we have a three-layer neural network where $z^i$ denotes the input of layer $i$, $\tanh(z)$ is the middle layer, and $L(z^{i-2})$ is the loss function. We then have 
$\frac{\partial L(z^{i-2})}{\partial z^{i-2}} = \frac{\partial L}{\partial z^{i + 1}} \frac{\partial \tanh(z^{i - 1})}{\partial z^{i-1}} \frac{\partial z^{i-1}}{\partial z^{i-2}}$. 
If the $\tanh$ layer is saturated, the gradient propagated to the layers below will be small, and no learning based on back propagation will occur.

If no parameter update is required for the bottom linear layer, we will also not run into the issue of saturated activation functions. However, in our experiments, we find that parameter update is necessary for the bottom linear layer because of covariate shift~\cite{SugiyamaKM07}, which is caused by discrepancy in the distribution between the source and target domains. If the feature distribution differs between domains, updating parameters is a straightforward approach to adapt the model for new domains. 

Although the two-layer neural network is capable of recognising target NE types, it has still two drawbacks. First, unlike a CRF, it doesn't include a label transition matrix. Second, the two-layer neural network has limited capacity if the domain discrepancy is large. If we rewrite the two-layer architecture in a compact way, we obtain:
\begin{equation}
 p(y' | \mathbf{x}) = \sigma(y',\mathbf{x}_i ;  \mathbf{W^t}\mathbf{W^s}).
\end{equation}
As the equation suggests, if we minimize the negative log likelihood, the loss function is not convex. Thus, we could land in a non-optimal local minimum using online learning. The pre-trained parameter matrix $\mathbf{W^s}$ imposes a special constraint that the computed scores for each target type are a weighted combination of updated source type scores. If a target type shares nothing in common with source types, the pre-trained $\mathbf{W^s}$ does more harm than good.

In the last step, we initialise the model parameters of a linear-chain CRF for $f(y_l, \mathbf{x})$ using the model parameters from the previous step. Based on the architecture of the NN model, we can collapse the two linear transformations into one by:
\begin{equation}
\label{eq:init}
\mathbf{W}^f =  \mathbf{W^t} \mathbf{W^s},
\end{equation}
while initialising the other parameters of the CRF to zero.  After this transformation, each initialised parameter vector $\mathbf{W}^f_{.y}$ is a weighted linear combination of the updated parameter vectors of the source types. Compared to the second step, the loss function we have now is convex because it is exactly a linear-chain CRF. Our previous steps have provided guided initialization of the parameters by incorporating source domain knowledge. The model also  has significantly more freedom to adapt itself to the target types. In other words, collapsing the two matrices simplifies the learning task and removes the constraints imposed by the pre-trained $\mathbf{W^s}$.

Because the tokens of the class \class{O} are generally several orders of magnitude more frequent than the tokens of the NE types, and also because of covariate shift, we found that the predictions of the NN models are biased towards the class \class{O} (i.e.\ a non-NE). As a result, the parameters of each NE type will always include or be dominated by the parameters of \class{O} after initialisation. To ameliorate this effect, we renormalise $\mathbf{W}^t$ before applying the transformation, as in \eqnref{eq:init}.  We do not include the parameters of the source class \class{O} when we initialise parameters of the NE types, while copying the parameters of the source class \class{O} to the target class \class{O}. In particular, let $o$ be the index of source domain class \class{O}. For each parameter vector $\mathbf{W}^t_{i*}$ of NE type, we set $W^t_{io} = 0$. For the parameter vector for the target class \class{O}, we set only the element corresponding to the weight between source type \class{O} and target class \class{O} to 1, and other elements to 0.

Finally, we fine-tune the model over the target domain by maximising log likelihood. The training objective is convex, and thus the local optimum is also the global optimum. If we fully train the model, we will achieve the same model as if we trained from scratch over only the target domain. As the knowledge of the source domain is hidden in the initial weights, we want to keep the initial weights as long as they contribute to the predictive task. Therefore, we apply AdaGrad~\cite{rizzo2012nerd} with early stopping based on development data, so that the knowledge of the source domain is preserved as much as possible.

\begin{figure*}[t]
	\begin{subfigure}{.33\textwidth}
		\centering
		\includegraphics[scale=.31]{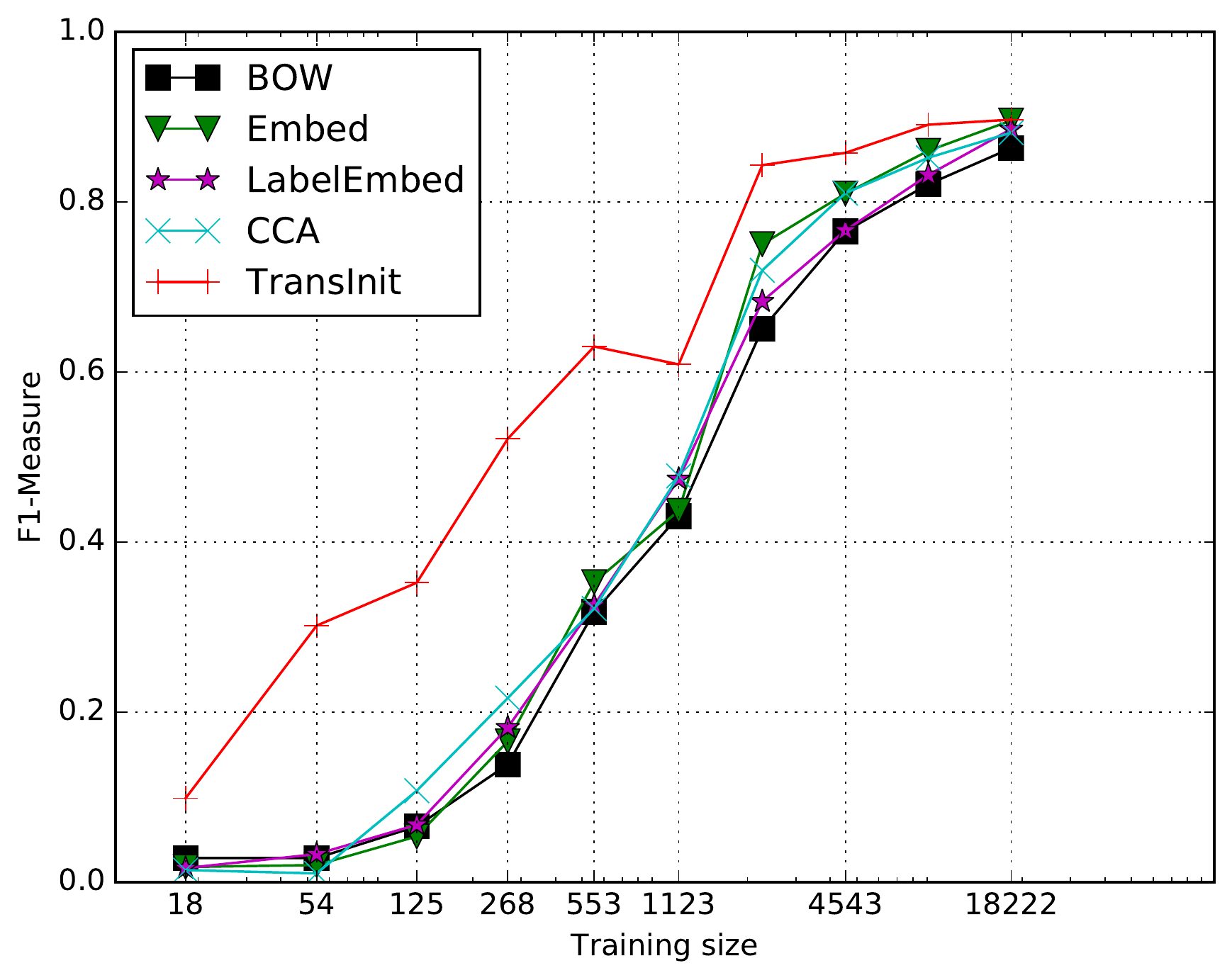}
		\caption{Target: \Ib, Source: \BBN}
		\label{fig:macro_bbn_i2b2_novel}
	\end{subfigure}
	\begin{subfigure}{.33\textwidth}
  	\centering
  	\includegraphics[scale=.31]{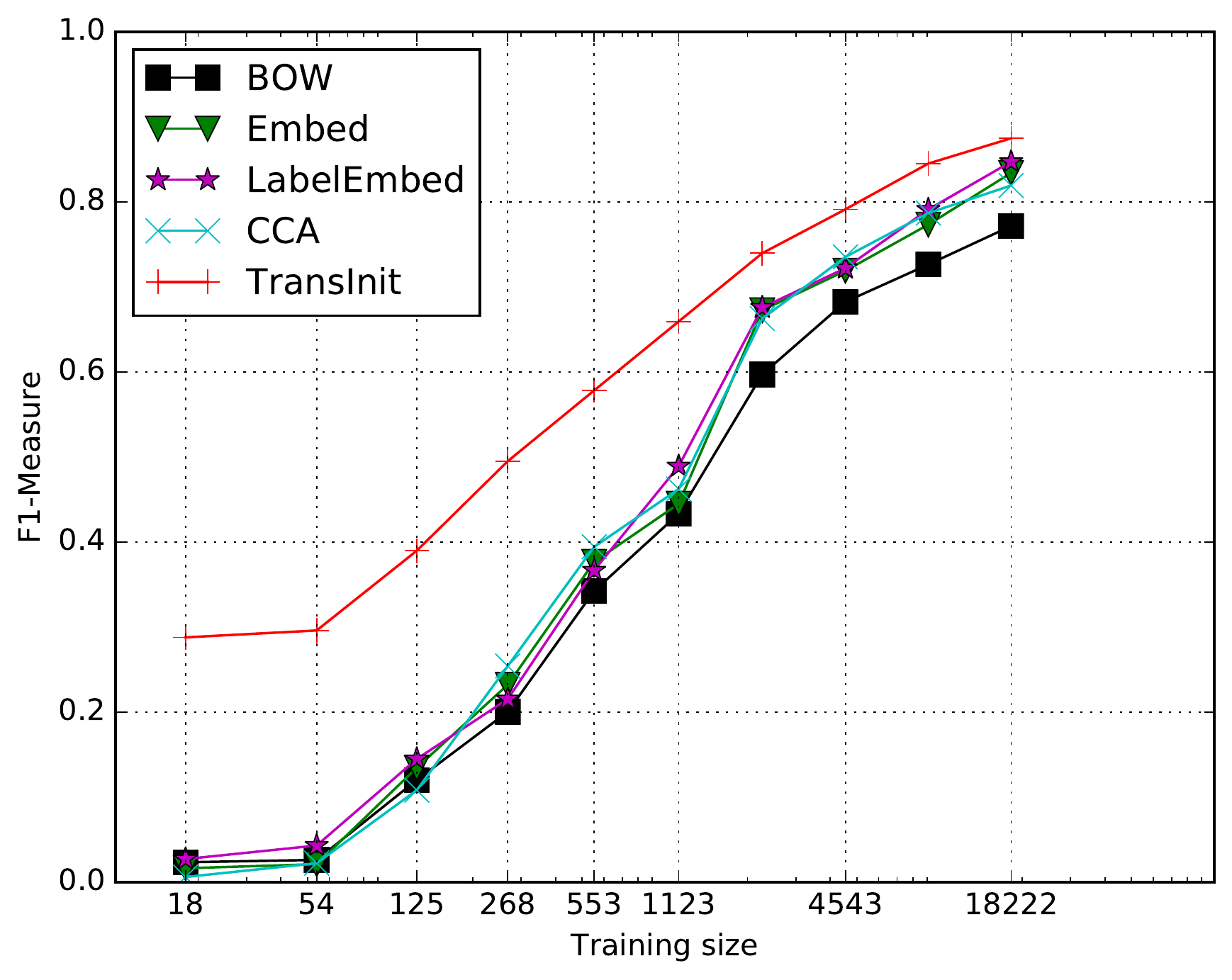}
  	\caption{Target: \Ib, Source: \Conll}
  	\label{fig:macro_conll_i2b2_novel}
	\end{subfigure}
	\begin{subfigure}{.33\textwidth}
		\centering
		\includegraphics[scale=.31]{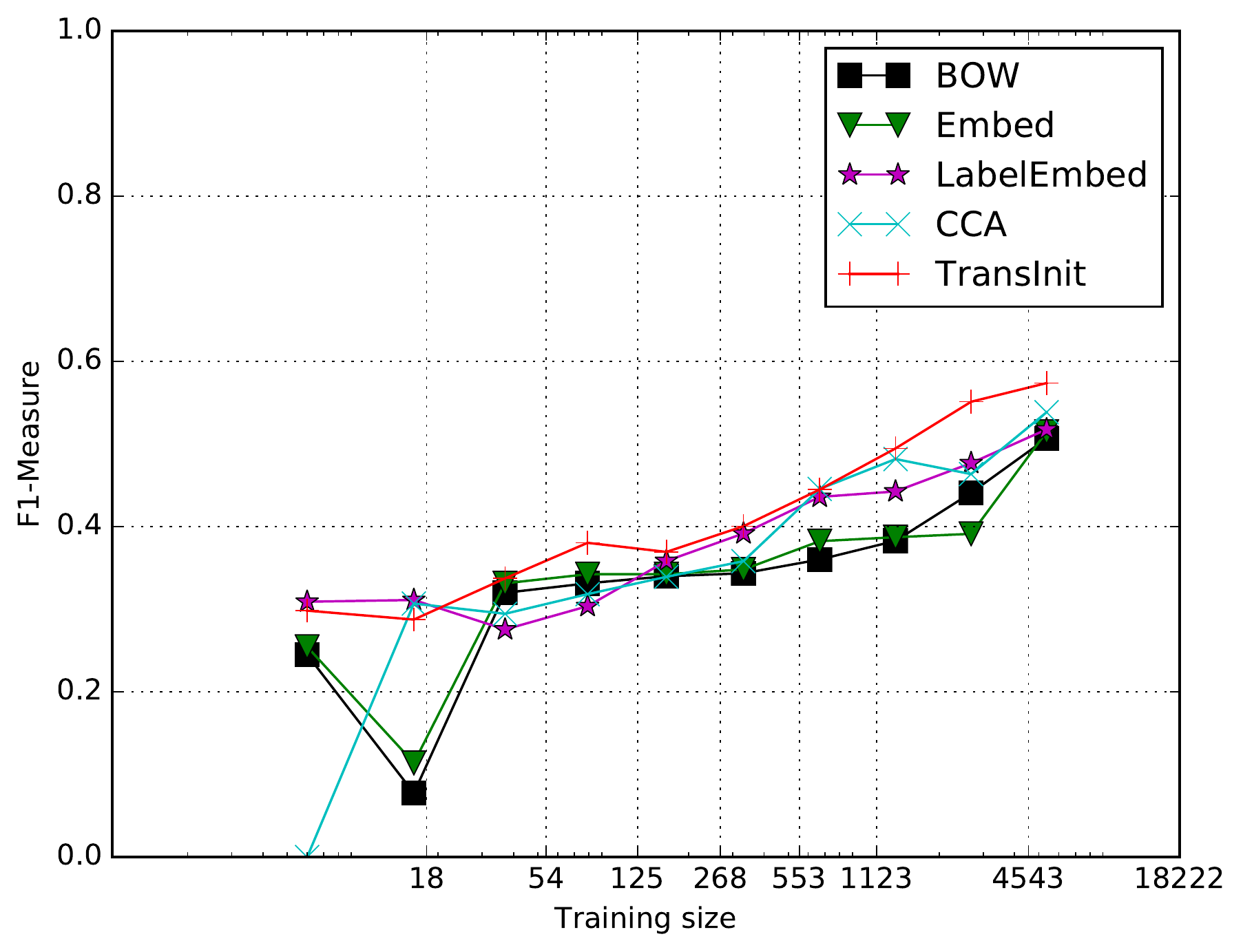}
		\caption{Target: \CADEC, Source: \Conll}
		\label{fig:macro_conll_cadec_novel}
	\end{subfigure}
	\caption{Macro-averaged F1 results across all novel classes on different source/target domain combinations}
	\label{fig:novel_types}
\end{figure*}

\section{Experimental Setup}
\label{sec:exp}

\subsection{Datasets}

We use \CADEC \cite{Karimi:2015} and \Ib \cite{abacha:2011} as target corpora with the standard training and test splits.
From each training set, we hold out 10\% as the development set. As source corpora, we adopt \Conll \cite{tjongkimsang2003conll} and \BBN \cite{Weischedel2005}.

In order to test the impact of the target domain training data size on results, we split the training set of \CADEC and \Ib into 10 partitions based on a log scale, and created 10 successively larger training sets by merging these partitions from smallest to largest (with the final merge resulting in the full training set). For all methods, we report the macro-averaged F1 over only the NE classes that are novel to the target domain.

\subsection{Baselines}
We compare our methods with the following two in-domain baselines, one cross-domain data-based method, and three cross-domain transfer-based benchmark methods.

\paragraph{BOW:} an in-domain linear-chain CRF with handcrafted features, from~\newcite{QuFZHSB15}.

\paragraph{Embed:} an in-domain linear-chain CRF with handcrafted features and pre-trained word embeddings, from~\newcite{QuFZHSB15}.

\paragraph{LabelEmbed:} take the labels in the source and target domains, and determine the alignment based on the similarity between the pre-trained embeddings for each label.

\paragraph{CCA:} the method of~\newcite{Kim:2015}, where a one-to-one mapping is generated between source and target NE classes using CCA and $k$-NN (see \secref{sec:relwork}).

\paragraph{\transDeep:} A three-layer deep CRF. The bottom layer is a linear layer initialised with $\mathbf{W}^s$ from the source domain-trained CRF. The middle layer is a hard $\tanh$ function~\cite{collobert2011natural}. The top layer is a linear-chain CRF with all parameters initialised to zero.

\paragraph{TwoLayerCRF:} A two-layer CRF. The bottom layer is a linear layer initialised with $\mathbf{W}^s$ from the source domain-trained CRF. The top layer is a linear-chain CRF with all parameters initialised to zero.\\

We compare our method with one variation, which is to freeze the parameters of the bottom linear layer and update only the parameters of the LR classifier while learning the correlation between the source and target types.

\subsection{Experimental Results}

\figref{fig:novel_types} shows the macro-averaged F1 of novel types between our method \transInit and the three baselines on all target corpora. The evaluation results on \CADEC with \BBN as the source corpus are not reported here because \BBN contains all types of \CADEC. From the figure we can see that \transInit outperforms all other methods with a wide margin on \Ib. When \Conll is taken as the source corpus, despite not sharing any NE types with \Ib, several target types are subclasses of source types: \class{DOCTOR} and \class{PATIENT} w.r.t.\ \class{PERSON},  and \class{HOSPITAL} w.r.t.\ \class{ORGANIZATION}. 

In order to verify if \transInit is able to capture semantic relatedness between source and target NE types, we inspected the parameter matrix $\mathbf{W}^t$ of the LR classifier in the step of learning type correlations. The corresponding elements in $\mathbf{W}^t$ indeed receive much higher values than the semantically-unrelated NE type pairs. When less than 300 target training sentences are used, these automatically discovered positive correlations directly lead to 10 times higher F1 scores for these types than the baseline \embed, which does not have a transfer learning step. Since \transInit is able to transfer the knowledge of multiple source types to related target types, this advantage leads to more than 10\% improvement in terms of F1 score on these types compared with \typeAlign, given merely 268 training sentences in \Ib. We also observe that, in case of few target training examples, \typeAlign is more robust than \cca if the correlation of types can be inferred from their names. 

We study the effects of transferring a large number of source types to target types by using \BBN, which has 64 types.  Here, the novel types of \Ib w.r.t.\ \BBN are \class{DOCTOR}, \class{PATIENT}, \class{HOSPITAL}, \class{PHONE}, and \class{ID}. For these types, \transInit successfully recognises \class{PERSON} as the most related type to \class{DOCTOR}, as well as \class{CARDINAL} as the most related type to \class{ID}. In contrast, \cca often fails to identify meaningful type alignments, especially for small training data sizes.

\CADEC is definitely the most challenging task when trained on \Conll, because there is no semantic connection between two of the target NE types (\class{DRUG} and \class{DISEASE}) and any of the source NE types. In this case, the baseline \typeAlign achieves competitive results with \transInit. This suggests that the class names reflect semantic correlations between source and target types, and there are not many shared textual patterns between any pair of source and target NE types in the respective datasets.
\begin{figure}[htb]
		\centering
		\includegraphics[scale=.35]{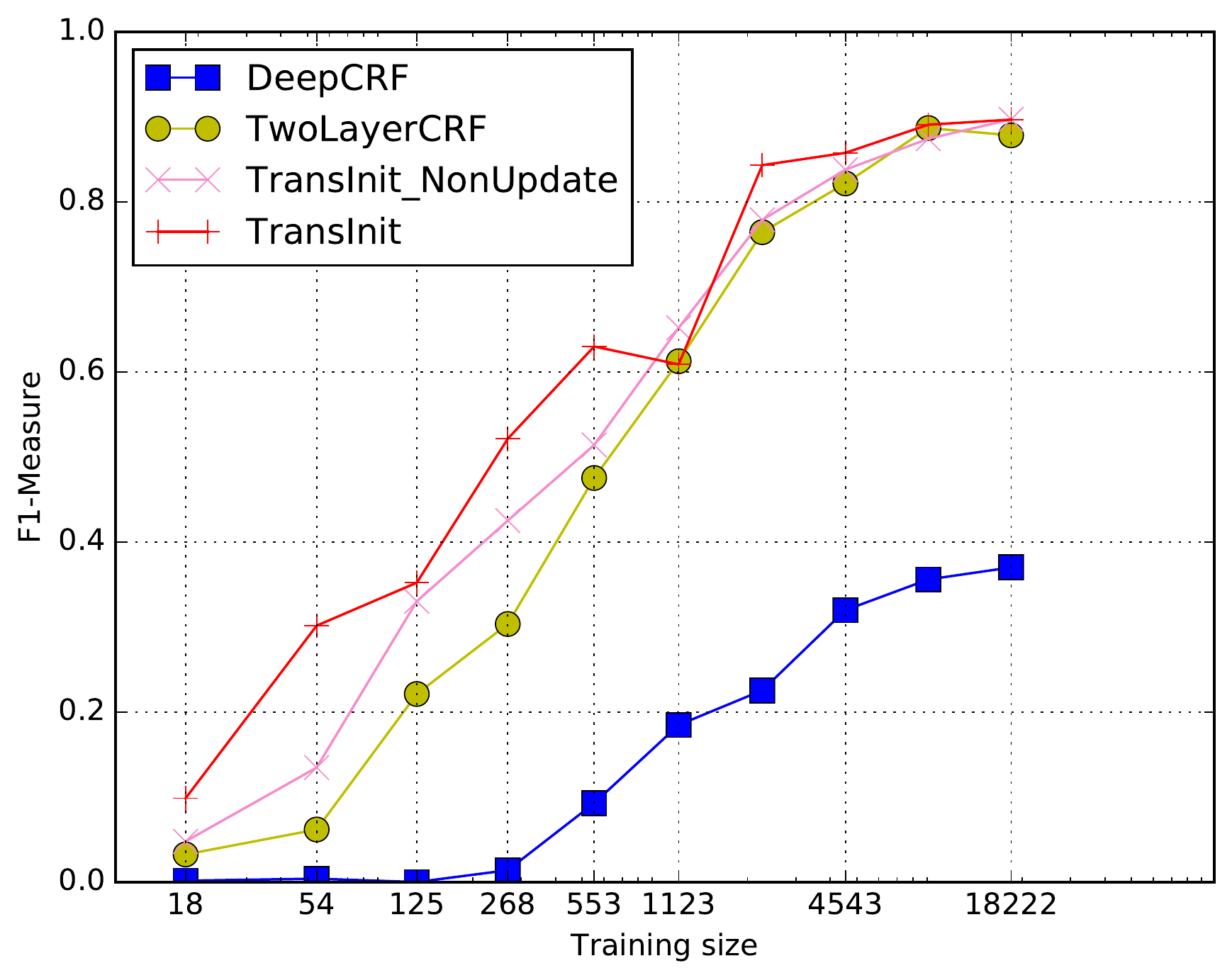}
			\caption{Difficulty of Transfer. The source model is trained on \BBN.}
	\label{fig:difficulty_transfer}
\end{figure}

Even with a complex model such as a neural network, the transfer of knowledge from the source types to the target types is not an easy task. \figref{fig:difficulty_transfer} shows that with a three-layer neural network, the whole model performs poorly. This is due to the fact that the hard $\tanh$ layer suffers from saturated function values. We inspected the values of the output hidden units computed by $\mathbf{W}^s \mathbf{x}$ on a random sample of target training examples before training on the target corpora. Most values are either highly positive or negative, which  is challenging for online learning algorithms. This is due to the fact that these hidden units are unnormalised probabilities produced by the source domain classifier. Therefore, removing the hidden non-linear-layer layer leads to a dramatic performance improvement.
Moreover, \figref{fig:difficulty_transfer} also shows that further performance improvement is achieved by reducing the two-layer architecture into a linear chain CRF.  And updating the hidden layers leads to up to 27\% higher F1 scores than not updating them in the second step of \transInit, which indicates that the neural networks need to update lower-level features to overcome the covariate shift problem.

\section{Conclusion}
\label{sec:conclusion}

We have proposed \transInit, a transfer learning-based method that supports the training of NER models across datasets where there are mismatches in domain and also possibly the label set. Our method was shown to achieve up to 160\% improvement in F1 over competitive baselines, based on a handful of in-domain training instances.


\section*{Acknowledgments}

This research was supported by NICTA, funded by the Australian Government through the Department of Communications and the Australian Research Council through the ICT Centre of Excellence Program. 

\bibliographystyle{emnlp2016}
\bibliography{biblio}

\end{document}